\documentclass{article}
\pdfoutput=1
\usepackage{spconf}

\usepackage[english]{babel}%
\usepackage[utf8]{inputenc} %
\usepackage{csquotes}%
\usepackage[backend=biber, 
            style=numeric,
            isbn=false,
            url=false,
            doi=false]{biblatex}
\AtEveryBibitem{%
  \clearfield{note}%
}

\usepackage{xspace} %
\usepackage{bm,bbm}
\usepackage{mathtools} %
\usepackage{amssymb}
\DeclareMathAlphabet{\mathcal}{OMS}{cmsy}{m}{n}
\usepackage[style=base,labelfont=bf]{caption}
\usepackage{subcaption}
\usepackage{booktabs} %
\usepackage{tabularx} %
\usepackage{multirow}
\ifpdf
   \usepackage{graphicx}
\else
   \usepackage{graphicx}
\fi
\usepackage[T1]{fontenc}    %
\usepackage{xcolor}
\usepackage{hyperref}       %
\usepackage{placeins}
\usepackage{url}            %
\usepackage{booktabs}       %
\usepackage{amsfonts}       %
\usepackage{nicefrac}       %
\usepackage{microtype}      %
\usepackage{tikz}
\usepackage[inline]{enumitem}
\usepackage{etoolbox}
\robustify{\subref}
\usepackage{breqn}
\newcommand{\bmu}[1]{\mathbf{#1}}           %
\newcommand{\xy}{\bmu{u}}

\newcommand{\CWT}{\ensuremath{{\mathbb{C}}\mathrm{WT}}\xspace} %
\newcommand{\DTCWT}{{\ensuremath{\mathrm{DT}\CWT}}\xspace}

\newcommand{\x}{\times}                     %

\newcommand{\conv}{\ast}

\newcommand{\F}[1]{\ensuremath{\mathrm{#1}}\xspace}

\newcommand{\norm}[1]{\left\lVert #1 \right\rVert}

\newcommand{\reals}[1][]{\ensuremath{{\mathbb{R}}^{#1}}\xspace}

\newcommand{\cnnfilt}[4]{h^{(#1)}_{#2}\left(#3, #4 \right)}

\newcommand{\cnnlact}[4]{#1^{(#2)}\left(#3, #4\right)}

\newif\ifcuboidshade
\newif\ifcuboidemphedge
\newif\ifcuboiddrawxdims
\newif\ifcuboiddrawydims
\newif\ifcuboiddrawzdims

\tikzset{
  cuboid/.is family,
  cuboid,
  shiftx/.initial=0,
  shifty/.initial=0,
  shiftz/.initial=0,
  dimx/.initial=3,
  dimy/.initial=3,
  dimz/.initial=3,
  scale/.initial=1,
  densityx/.initial=1,
  densityy/.initial=1,
  densityz/.initial=1,
  rotation/.initial=0,
  anglex/.initial=0,
  angley/.initial=90,
  anglez/.initial=225,
  scalex/.initial=1,
  scaley/.initial=1,
  scalez/.initial=0.5,
  front/.style={draw=black,fill=white},
  top/.style={draw=black,fill=white},
  right/.style={draw=black,fill=white},
  shade/.is if=cuboidshade,
  shadecolordark/.initial=black,
  shadecolorlight/.initial=white,
  shadeopacity/.initial=0.15,
  shadesamples/.initial=16,
  emphedge/.is if=cuboidemphedge,
  emphstyle/.style={thick},
  drawzdims/.is if=cuboiddrawzdims,
  dimzval/.initial=C,
  drawxdims/.is if=cuboiddrawxdims,
  dimxval/.initial=W,
  drawydims/.is if=cuboiddrawydims,
  dimyval/.initial=H,
}

\newcommand{\tikzcuboidkey}[1]{\pgfkeysvalueof{/tikz/cuboid/#1}}

\newcommand{\tikzcuboid}[1]{
    \tikzset{cuboid,#1} %
  \pgfmathsetlengthmacro{\vectorxx}{\tikzcuboidkey{scalex}*cos(\tikzcuboidkey{anglex})*28.452756}
  \pgfmathsetlengthmacro{\vectorxy}{\tikzcuboidkey{scalex}*sin(\tikzcuboidkey{anglex})*28.452756}
  \pgfmathsetlengthmacro{\vectoryx}{\tikzcuboidkey{scaley}*cos(\tikzcuboidkey{angley})*28.452756}
  \pgfmathsetlengthmacro{\vectoryy}{\tikzcuboidkey{scaley}*sin(\tikzcuboidkey{angley})*28.452756}
  \pgfmathsetlengthmacro{\vectorzx}{\tikzcuboidkey{scalez}*cos(\tikzcuboidkey{anglez})*28.452756}
  \pgfmathsetlengthmacro{\vectorzy}{\tikzcuboidkey{scalez}*sin(\tikzcuboidkey{anglez})*28.452756}
  \begin{scope}[
    xshift=\tikzcuboidkey{shiftx}, 
    yshift=\tikzcuboidkey{shifty}, 
    scale=\tikzcuboidkey{scale}, 
    rotate=\tikzcuboidkey{rotation}, 
    x={(\vectorxx,\vectorxy)}, 
    y={(\vectoryx,\vectoryy)}, 
    z={(\vectorzx,\vectorzy)}]

    \pgfmathsetmacro{\steppingx}{1/\tikzcuboidkey{densityx}}
    \pgfmathsetmacro{\steppingy}{1/\tikzcuboidkey{densityy}}
    \pgfmathsetmacro{\steppingz}{1/\tikzcuboidkey{densityz}}
    \newcommand{\dimx}{\tikzcuboidkey{dimx}}
    \newcommand{\dimy}{\tikzcuboidkey{dimy}}
    \newcommand{\dimz}{\tikzcuboidkey{dimz}}
    \newcommand{\shiftz}{\tikzcuboidkey{shiftz}}
    \pgfmathsetmacro{\secondx}{2*\steppingx}
    \pgfmathsetmacro{\secondy}{2*\steppingy}
    \pgfmathsetmacro{\secondz}{2*\steppingz}
    \foreach \x in {\steppingx,\secondx,...,\dimx} { 
      \foreach \y in {\steppingy,\secondy,...,\dimy} {   
        \pgfmathsetmacro{\lowx}{(\x-\steppingx)}
        \pgfmathsetmacro{\lowy}{(\y-\steppingy)}
        \filldraw[cuboid/front] (\lowx,\lowy,0.5*\dimz+\shiftz) -- (\lowx,\y,0.5*\dimz+\shiftz) -- (\x,\y,0.5*\dimz+\shiftz) -- (\x,\lowy,0.5*\dimz+\shiftz) -- cycle;
      }
    }
    \foreach \x in {\steppingx,\secondx,...,\dimx} { 
      \foreach \z in {\steppingz,\secondz,...,\dimz} {   
        \pgfmathsetmacro{\lowx}{(\x-\steppingx)}
        \pgfmathsetmacro{\lowz}{(\z-\steppingz-0.5*\dimz+\shiftz)}
        \pgfmathsetmacro{\highz}{(\z-0.5*\dimz+\shiftz)}
        \filldraw[cuboid/top] (\lowx,\dimy,\lowz) -- (\lowx,\dimy,\highz) -- (\x,\dimy,\highz) -- (\x,\dimy,\lowz) -- cycle;
      }
    }
    \foreach \y in {\steppingy,\secondy,...,\dimy} { 
      \foreach \z in {\steppingz,\secondz,...,\dimz} {
        \pgfmathsetmacro{\lowy}{(\y-\steppingy)}
        \pgfmathsetmacro{\lowz}{(\z-\steppingz-0.5*\dimz+\shiftz)}
        \pgfmathsetmacro{\highz}{(\z-0.5*\dimz+\shiftz)}
        \filldraw[cuboid/right] (\dimx,\lowy,\lowz) -- (\dimx,\lowy,\highz) -- (\dimx,\y,\highz) -- (\dimx,\y,\lowz) -- cycle;
      }
    }
    \ifcuboidemphedge
      \draw[cuboid/emphstyle] (0,\dimy,-0.5*\dimz+\shiftz) -- (\dimx,\dimy,-0.5*\dimz+\shiftz) -- (\dimx,\dimy,0.5*\dimz+\shiftz) -- (0,\dimy,0.5*\dimz+\shiftz) -- cycle;%
      \draw[cuboid/emphstyle] (0,\dimy,0.5*\dimz+\shiftz) -- (0,0,0.5*\dimz+\shiftz) -- (\dimx,0,0.5*\dimz+\shiftz) -- (\dimx,\dimy,0.5*\dimz+\shiftz);%
      \draw[cuboid/emphstyle] (\dimx,\dimy,-0.5*\dimz+\shiftz) -- (\dimx,0,-0.5*\dimz+\shiftz) -- (\dimx,0,0.5*\dimz+\shiftz);%
    \fi
    \ifcuboiddrawxdims
      \draw[<->] (0, -0.5, 0.5*\dimz+\shiftz) -- (\dimx, -0.5, 0.5*\dimz+\shiftz) node[below,midway] {$\tikzcuboidkey{dimxval}$};
    \fi
    \ifcuboiddrawydims
      \draw[<->] (-0.5, 0, 0.5*\dimz+\shiftz) -- (-0.5, \dimy, 0.5*\dimz+\shiftz) node[left,midway] {$\tikzcuboidkey{dimyval}$};
    \fi
    \ifcuboiddrawzdims
      \draw[<->] (\dimx, -0.5, -0.5*\dimz+\shiftz) -- (\dimx, -0.5, 0.5*\dimz+\shiftz) node[below right,midway] {$\tikzcuboidkey{dimzval}$};
    \fi

    \ifcuboidshade
      \pgfmathsetmacro{\cstepx}{\dimx/\tikzcuboidkey{shadesamples}}
      \pgfmathsetmacro{\cstepy}{\dimy/\tikzcuboidkey{shadesamples}}
      \pgfmathsetmacro{\cstepz}{\dimz/\tikzcuboidkey{shadesamples}}
      \foreach \s in {1,...,\tikzcuboidkey{shadesamples}} {   
        \pgfmathsetmacro{\lows}{\s-1}
        \pgfmathsetmacro{\cpercent}{(\lows)/(\tikzcuboidkey{shadesamples}-1)*100}
        \fill[opacity=\tikzcuboidkey{shadeopacity},
              color=\tikzcuboidkey{shadecolorlight}!\cpercent!\tikzcuboidkey{shadecolordark}] 
            (0,\s*\cstepy,0.5*\dimz+\shiftz) -- (\s*\cstepx,\s*\cstepy,0.5*\dimz+\shiftz) -- (\s*\cstepx,0,0.5*\dimz+\shiftz) 
              -- (\lows*\cstepx,0,0.5*\dimz+\shiftz) -- (\lows*\cstepx,\lows*\cstepy,0.5*\dimz+\shiftz) -- (0,\lows*\cstepy,0.5*\dimz+\shiftz) -- cycle;
        \fill[opacity=\tikzcuboidkey{shadeopacity},
              color=\tikzcuboidkey{shadecolorlight}!\cpercent!\tikzcuboidkey{shadecolordark}] 
            (0,\dimy,\s*\cstepz-0.5*\dimz+\shiftz) -- (\s*\cstepx,\dimy,\s*\cstepz-0.5*\dimz+\shiftz) -- (\s*\cstepx,\dimy,-0.5*\dimz+\shiftz) 
              -- (\lows*\cstepx,\dimy,-0.5*\dimz+\shiftz) -- (\lows*\cstepx,\dimy,\lows*\cstepz-0.5*\dimz+\shiftz) -- (0,\dimy,\lows*\cstepz-0.5*\dimz+\shiftz) -- cycle;
        \fill[opacity=\tikzcuboidkey{shadeopacity},
              color=\tikzcuboidkey{shadecolorlight}!\cpercent!\tikzcuboidkey{shadecolordark}] 
            (\dimx,0,\s*\cstepz-0.5*\dimz+\shiftz) -- (\dimx,\s*\cstepy,\s*\cstepz-0.5*\dimz+\shiftz) -- (\dimx,\s*\cstepy,-0.5*\dimz+\shiftz) 
              -- (\dimx,\lows*\cstepy,-0.5*\dimz+\shiftz) -- (\dimx,\lows*\cstepy,\lows*\cstepz-0.5*\dimz+\shiftz) -- (\dimx,0,\lows*\cstepz-0.5*\dimz+\shiftz) -- cycle;
      }
    \fi 

  \end{scope}
}

\usepackage{array}
\newcommand{\PreserveBackslash}[1]{\let\temp=\\#1\let\\=\temp}
\newcolumntype{C}[1]{>{\PreserveBackslash\centering}p{#1}}
\newcolumntype{R}[1]{>{\PreserveBackslash\raggedleft}p{#1}}
\newcolumntype{L}[1]{>{\PreserveBackslash\raggedright}p{#1}}

\title{A Learnable ScatterNet: Locally Invariant Convolutional Layers}

\name{Fergal Cotter \qquad Nick Kingsbury}
\address{Signal Processing and Communications Group, Engineering Department\\
         University of Cambridge, U.K.\\
         \texttt{fbc23@cam.ac.uk}, \texttt{ngk@eng.cam.ac.uk}}

\begin{document}
\ninept

\maketitle

\begin{abstract}
  In this paper we explore tying together the ideas from Scattering Transforms
  and Convolutional Neural Networks (CNN) for Image Analysis by proposing a learnable
  ScatterNet. Previous attempts at tying them together in hybrid 
  networks have tended to keep the two parts separate, with the ScatterNet forming a fixed
  front end and a CNN forming a learned backend. We instead look at adding learning
  between scattering orders, as well as adding learned layers before the
  ScatterNet. We do this by breaking down the scattering orders into single
  convolutional-like layers we call `locally invariant' layers, and adding a learned 
  mixing term to this layer. Our experiments show that these locally invariant
  layers can improve accuracy when added to either a CNN or a ScatterNet.
  We also discover some surprising results in that the ScatterNet may be best
  positioned after one or more layers of learning rather than at the front of a neural
  network.
\end{abstract}


\section{Introduction}

In image understanding tasks such as classification, recognition and segmentation,
Convolutional Neural Networks (CNNs) have now become the \emph{de facto} and the state
of the art model. Since they proved their worth in 2012 by winning the ImageNet
Large Scale Visual Recognition Competition (ILSVRC) \cite{russakovsky_imagenet_2014}
with the AlexNetwork \cite{krizhevsky_imagenet_2012}, they have been fine tuned and developed into very
powerful and flexible models. Most of this development has been in changing the
architecture, such as the Residual Network\cite{he_deep_2016}, the Inception
Network \cite{szegedy_going_2015} and the DenseNet \cite{huang_densely_2017}.
However one of the key building blocks of CNNs, the convolutional filter bank,
has seen less development and in today's models they are not too dissimilar to
what they were in 2012. %
We believe there is still much to explore in the way convolutional 
filters are built and learned. 

Current layers are randomly initialized and learned through backpropagation by
minimizing a custom loss function.
The properties and purpose of learned filters
past the first layer of a CNN are not well understood, a deeply unsatisfying
situation, particularly when we start to see problems in modern solutions such as significant redundancy
\cite{denton_exploiting_2014} and weakness to adversarial attacks
\cite{carlini_towards_2017}. 

The Scattering Transform by Mallat et.\ al.\ \cite{mallat_group_2012, bruna_invariant_2013} 
attempts to address the problems of poorly understood filtering layers 
by using predefined wavelet bases whose properties are well known. Using
this knowledge, Mallat derives bounds on the effect of 
noise and deformations to the input. This work inspires us, but the
fixed nature of ScatterNets has proved a limiting factor for them so far.

To combat this, \cite{oyallon_scaling_2017, oyallon_hybrid_2017,
singh_scatternet_2018} use ScatterNets as a front end for deep learning tasks,
calling them Hybrid ScatterNets. These have had some success, but
in \cite{cotter_visualizing_2017} we visualized what features a ScatterNet extracted
and showed that they were ripple-like and very dissimilar from what a CNN would
extract, suggesting that some learning should be done between the ScatterNet
orders. 

To do this, we take inspiration from works like \cite{qiu_dcfnet:_2018,
jacobsen_dynamic_2017, worrall_harmonic_2017, cotter_deep_2018},
which decompose convolutional filters as a learned mixing of fixed harmonic
functions. However we now propose to build a
convolutional-like layer which is a learned mixing of the locally invariant
scattering terms from the original ScatterNet \cite{bruna_invariant_2013}.

In \autoref{sec:background} we briefly review convolutional layers and scattering
layers before introducing our learnable scattering layers in \autoref{sec:method}.
In \autoref{sec:implementation} we describe how we implement our proposed layer,
and present some experiments we have run in \autoref{sec:experiments} and then
draw conclusions about how these new ideas might improve neural networks in the
future.

\vspace{-5pt}
\section{Related Work}\label{sec:background}

\subsection{Convolutional Layers}\label{sec:conv_layer}

Let the output of a CNN at layer $l$ be:

$$ \cnnlact{x}{l}{c}{\xy}, \quad c\in \{0, \ldots C_l-1\}, \xy \in \reals[2]$$

where $c$ indexes the channel dimension , and $\xy$ is a vector of coordinates
for the spatial position. Of course, $\xy$ is typically sampled on a grid, but
following the style of \cite{qiu_dcfnet:_2018}, we keep it continuous to
more easily differentiate between the spatial and channel dimensions. A typical
convolutional layer in a standard CNN (ignoring the bias term) is:
\vspace{-15pt}
\begin{eqnarray} 
  \cnnlact{y}{l+1}{f}{\xy} &=& \sum_{c=0}^{C_l - 1}  x^{(l)}(c,\xy) \conv h^{(l)}_{f}(c, \xy)
    \label{eq:conv}\\
    \cnnlact{x}{l+1}{f}{\xy} & = & \sigma \left( \cnnlact{y}{l+1}{f}{\xy} \right) \label{eq:nonlin}
\end{eqnarray}

where $\cnnfilt{l}{f}{c}{\xy}$ is the $f$th filter of the $l$th layer (i.e. $f \in \{0,
\ldots, C_{l+1}-1 \}$) with $C_l$ different point spread functions. $\sigma$ is a non-linearity 
possibly combined with scaling such as batch normalization. The convolution
is done independently for each $c$ in the $C_l$ channels and the resulting outputs are
summed together to give one activation map. This is repeated $C_{l+1}$ times to
give $\left\{ \cnnlact{x}{l+1}{f}{\xy} \right\}_{f \in \{0, \ldots, C_{l+1}-1 \}, \xy \in \reals[2]}$

\subsection{Wavelets and Scattering Transforms}\label{sec:scatternet}
The 2-D wavelet transform is done by convolving the input with a mother wavelet
dilated by $2^j$ and rotated by $\theta$:

\begin{equation}
  \psi_{j, \theta}(\xy) = 2^{-j}\psi \left(2^{-j} R_{-\theta} \xy\right)
\end{equation}

where $R$ is the rotation matrix, $1 \leq j \leq J$ indexes the scale, and
$1 \leq k \leq K$ indexes $\theta$ to give $K$ angles between $0$ and $\pi$. We
copy notation from \cite{bruna_invariant_2013} and define $\lambda = (j, k)$ and
the set of all possible $\lambda$s is $\Lambda$ whose size is $|\Lambda | = JK$.
The wavelet transform, including lowpass, is then:
\begin{equation}
  Wx(c, \xy) = \left\{ x(c, \xy)\conv \phi_J(\xy), x(c, \xy)\conv \psi_\lambda (\xy) \right\}_{\lambda \in \Lambda}
\end{equation}

Taking the modulus of the wavelet coefficients removes the high frequency
oscillations of the output signal while preserving the energy of the
coefficients over the frequency band covered by $\psi_\lambda$. This is crucial
to ensure that the scattering energy is concentrated towards zero-frequency as
the scattering order increases, allowing sub-sampling.
We define the wavelet modulus propagator to be:
\begin{equation}
  \label{eq:wave_mod}
  \tilde{W}x(c, \xy) = \left\{ x(c, \xy) \conv \phi_{J}(\xy), |x(c, \xy) \conv \psi_\lambda (\xy) | \right\}_{\lambda \in \Lambda} 
\end{equation}

Let us call these modulus terms $U[\lambda] x = \lvert x \conv \psi_\lambda
\rvert$ and define a path as a sequence of $\lambda$s given by $p = \left(\lambda_1,
\lambda_2, \ldots \lambda_m \right)$. Further, define the modulus propagator
acting on a path $p$ by:
\begin{eqnarray}
  U[p]x & = & U[\lambda_m] \cdots U[\lambda_2]U[\lambda_1]x \label{eq:u_paths}\\
        & = & || \cdots | x\conv \psi_{\lambda_1} | \conv \psi_{\lambda_2} | \cdots \conv \psi_{\lambda_m} |
\end{eqnarray}

These descriptors are then averaged over the window $2^J$ by a scaled lowpass filter $\phi_J =
2^{-J}\phi(2^{-J}\xy)$ giving the `invariant' scattering coefficient
\begin{equation}
  S[p]x(\xy) = U[p]x \conv \phi_J(\xy)
\end{equation}

If we define $p + \lambda = (\lambda_1, \ldots, \lambda_m, \lambda)$ then we can
combine \autoref{eq:wave_mod} and \autoref{eq:u_paths} to give:
\begin{equation}
  \tilde{W} U[p] x = \left\{S[p]x, U[p+\lambda]x \right\}_{\lambda}
\end{equation}

Hence we iteratively apply $\tilde{W}$ to all of the propagated $U$ terms of
the previous layer to get the next order of scattering coefficients and
the new $U$ terms.

The resulting scattering coefficients have many nice properties, one of which is
stability to diffeomorphisms (such as shifts and warping). From
\cite{mallat_group_2012}, if $\mathcal{L}_\tau
x = x(\xy -\tau(\xy))$ is a diffeomorphism which is bounded with 
$\norm{\nabla \tau}_{\infty} \leq 1/2$, then there exists a $K_L > 0$ such
that:
\begin{equation}
  \norm{ S \mathcal{L}_{\tau}x  - S x} \leq K_L P H(\tau) \norm{x}
  \label{eq:stability}
\end{equation}

where $P = \F{length}(p)$ is the scattering order, and $H(\tau)$ is a function
of the size of the displacement, derivative and Hessian of $\tau$.

\section{Locally Invariant Layer} \label{sec:method}
The $U$ terms in \autoref{sec:scatternet} are often called `covariant' terms but
in this paper we will call them locally invariant, as they tend to be invariant up to a
scale $2^j$.  We propose to mix the locally invariant terms $U$ and the
lowpass terms $S$ with learned weights $a_{f,\lambda}$ and $b_f$. For example,
consider a first order ScatterNet, and let the input to it be $x^{(l)}$. Our
proposed output $y^{(l+1)}$ is then:
\begin{eqnarray} \label{eq:comb1}
  y^{(l+1)}(f, \xy) & = & \sum_{\lambda\in \Lambda} \sum_{c=0}^{C-1} |x^{(l)}(c, \xy)\conv \psi_\lambda(\xy)|\ a_{f, \lambda}(c) \nonumber \\
                    & + & \left(\sum_{c=0}^{C-1} x^{(l)}(c, \xy) \conv \phi_J(\xy)\right) b_f(c)
  \label{eq:layer}
\end{eqnarray}

Returning to \autoref{eq:wave_mod}, we define a new index variable $\gamma$
such that $\tilde{W}[\gamma]x = x\conv \phi_J$ for $\gamma = 1$ and
$\tilde{W}[\gamma]x = |x \conv \psi_{\lambda}|$ for $2 \leq \gamma \leq JK + 1$.
We do the same for the weights $a$, $b$ by defining $\tilde{a}_f = \left\{b_f,
a_{f, \lambda} \right\}_\lambda$ and $\tilde{a}_f[\gamma] = b_f$ if $\gamma = 1$
and $\tilde{a}_f[\gamma] = a_{f, \lambda}$ if $2 \leq \gamma \leq JK + 1$.
We further define $q = (c, \gamma) \in Q$ to combine the channel
and index terms.  This simplifies \autoref{eq:layer} to be:
\begin{eqnarray}
  z^{(l+1)}(q, \xy) & = & \tilde{W}x^{(l)}[q] = \tilde{W}x^{(l)}[\tilde{\lambda}](c, \xy) \label{eq:covariant}\\
  y^{(l+1)}(f, \xy) & = & \sum_{q \in Q} z^{(l+1)}(q, \xy) \tilde{a}_f(q) \label{eq:mixing}
\end{eqnarray}
or in matrix form with $A_{f,q} = \tilde{a}_f(q)$
\begin{equation}
  Y^{(l+1)}(\xy)  =  A Z^{(l+1)}(\xy) \label{eq:matrix}
\end{equation}

This is very similar to the standard convolutional layer from
\autoref{eq:conv}, except we have replaced the previous layer's $x$ with
intermediate coefficients $z$ (with $|Q| = (JK+1)C$ channels), and the
convolutions of \autoref{eq:conv} have been replaced by a matrix multiply
(which can also be seen as a $1\x 1$ convolutional layer). We can then apply
\autoref{eq:nonlin} to \autoref{eq:mixing} to get the next layer's output (or
equivalently, the next order scattering coefficients). 

\subsection{Properties}
The first thing to note is that with careful choice of $A$ and $\sigma$, we can
recover the original translation invariant ScatterNet
\cite{bruna_invariant_2013, oyallon_scaling_2017}. If $C_{l+1} = (JK+1)C_l$ 
and $A$ is the identity matrix $I_{C_{l+1}}$, we remove the mixing and then $y^{(l+1)} = \tilde{W}x$.

Further, if $\sigma = \F{ReLU}$ as is commonly the case in training CNNs, it has
no effect on the positive locally invariant terms $U$. It will affect the averaging terms
if the signal is not positive, but this can be dealt with by adding a channel
dependent bias term $\alpha_c$ to $x^{(l)}$ to ensure it is positive. This bias term
will not affect the propagated signals as $\int \alpha_c \psi_\lambda(\xy) d\xy =
0$. The bias can then be corrected by subtracting $\alpha_c \norm{\phi_J}_2$ from
the averaging terms after taking the ReLU, then $x^{(l+1)} = \tilde{W}x$.

This makes one layer of our system equivalent to a first order scattering
transform, giving $S_0$ and $U_1$ (invariant to input shifts of $2^1$). Repeating the
same process for the next layer again works, as we saw in \autoref{eq:u_paths},
giving $S_1$ and $U_2$ (invariant to shifts of $2^2$).  If we want to build
higher invariance, we can continue or simply average these outputs with an average pooling
layer.

Let us define the action of our layer on the scattering
coefficients to be $Vx$. We would like to find a bound on $\norm{V\mathcal{L}_\tau x -
V x}$. To do this, we note that the mixing is a linear operator and hence is
Lipschitz continuous. The authors in \cite{qiu_dcfnet:_2018} find constraints on the mixing
weights to make them non-expansive (i.e. Lipschitz constant 1).
Further, the ReLU is non-expansive meaning the combination of the two is
also non-expansive, so $\norm{V\mathcal{L}_\tau x - V x} \leq \norm{S\mathcal{L}_\tau
x - Sx}$, and \autoref{eq:stability} holds.

\begin{figure*}[ht!]
  \centering
  \small
  \resizebox{\textwidth}{!} {
  \begin{tikzpicture}[%
    path image/.style={
      path picture={
        \node at (path picture bounding box.center) {
          \includegraphics[height=2.0cm]{#1}
        };
      }
    }, 
    path pic/.style={
      path picture={
        \node at (path picture bounding box.center) {
          \includegraphics[height=1.2cm]{#1}
        };
      }
    }, 
    path pic2/.style={
      path picture={
        \node at (path picture bounding box.center) {
          \includegraphics[height=0.8cm]{#1}
        };
      }
    }, 
    scale=0.6]

    \tikzcuboid{
    shiftx=-1.5cm,
    shifty=-2.5cm,
    scale=0.5,
    anglex=0, 
    angley=90, 
    anglez=230,
    dimx=3, 
    dimy=3, 
    dimz=6,
    densityx=1, 
    densityy=1, 
    densityz=1,
    shade=false,
    emphedge=true,
    shadeopacity=0,
    emphstyle/.style={rounded corners=0.2pt,line width=0.3mm},
    front/.style={draw=blue!50!white,fill=blue!50!white},%
    right/.style={draw=blue!50!white,fill=blue!50!white},%
    top/.style={draw=blue!50!white,fill=blue!50!white},%
    drawxdims=true,
    dimxval=W,
    drawydims=true,
    dimyval=H,
    drawzdims=true,
    dimzval=C_l,
    }
    \draw (0, .3, 0) node {\large{$x^{(l)}$}};
    \draw [path image=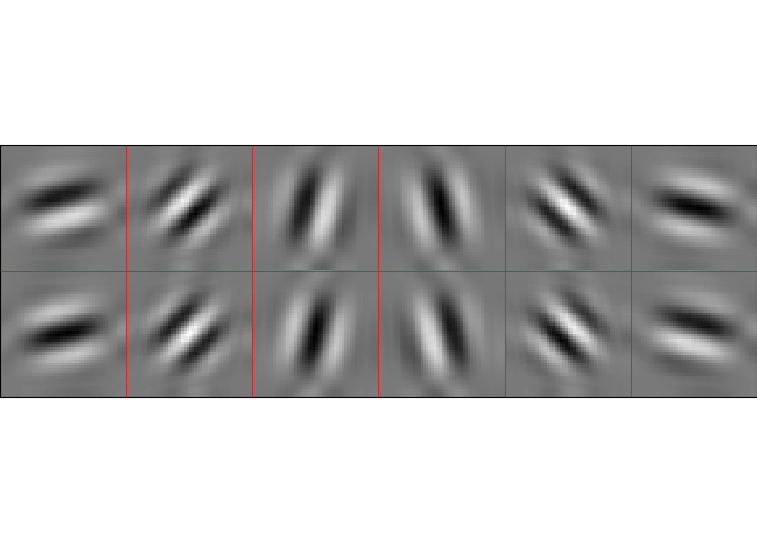, draw=black] (1.5,1.5,0) rectangle (6,0,0);
    \draw (3.75, 1.9, 0) node {\large{$\psi_{j, \theta}$}};
    \draw (1.5,0,0) -- (3,-1.7,0);
    \draw (6,0,0) -- (3.3,-1.7,0);
    \draw [path pic2=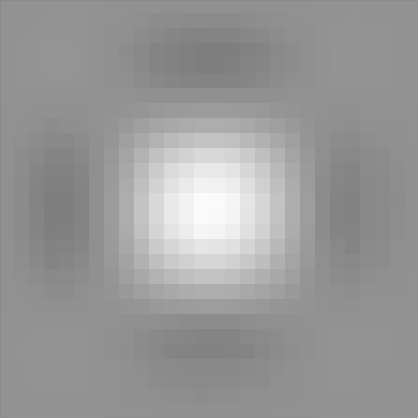, draw=black] (2.5,-2.5,10) rectangle (3.5,-1.5,10);
    \draw (3, -1.1, 10) node {\large{$\phi_{j}$}};
    \draw (3.5,-1.5,10) -- (3.5,-1.7,8);

    \draw (2.5, -1.5, 3) node {\Large{$\conv$}};
    \draw [->, fill=gray!30,ultra thick] (4, -1.5, 0) -- (5, -1.5, 0);
    \draw [->, fill=gray!30,ultra thick] (4.5, -1.5, 9) -- (5.85, -1.5, 11);

    \tikzcuboid{
    shiftx=3cm,
    shifty=-1.7cm,
    shiftz=0,
    scale=0.3,
    dimx=1, dimy=1, dimz=4,
    densityx=2, densityy=2, densityz=2,
    drawxdims=false,
    drawydims=false,
    drawzdims=true,
    dimzval=12,
    front/.style={draw=yellow!70!white,fill=yellow!70!white},%
    right/.style={draw=yellow!70!white,fill=yellow!70!white},%
    top/.style={draw=yellow!70!white,fill=yellow!70!white},%
    }
    \tikzcuboid{
    shiftz=8/0.3,
    scale=0.3,
    dimx=1, dimy=1, dimz=1,
    densityx=2, densityy=2, densityz=2,
    drawxdims=false,
    drawydims=false,
    drawzdims=false,
    }

    \tikzcuboid{
    shiftx=6cm,
    shifty=-2.0cm,
    shiftz=0.8cm,
    scale=0.5,
    dimx=2, dimy=2, dimz=4,
    densityx=4, densityy=4, densityz=2,
    drawzdims=true,
    dimzval=C_l,
    front/.style={draw=blue!50!white,fill=blue!50!white},%
    right/.style={draw=blue!50!white,fill=blue!50!white},%
    top/.style={draw=blue!50!white,fill=blue!50!white},%
    }

    \tikzcuboid{
    shiftx=6cm,
    shifty=-2cm,
    shiftz=0cm,
    scale=0.5,
    dimx=2, dimy=2, dimz=24,
    densityx=2, densityy=2, densityz=2,
    drawxdims=true,
    dimxval=\frac{W}{2},
    drawydims=true,
    dimyval=\frac{H}{2},
    drawzdims=true,
    dimzval=12C_l,
    front/.style={draw=blue!50!white,fill=blue!50!white},%
    right/.style={draw=blue!50!white,fill=blue!50!white},%
    top/.style={draw=blue!50!white,fill=blue!50!white},%
    }

    \draw [->, fill=gray!30,ultra thick] (8.5, -1.5, 0) -- (10.5, -1.5, 0) node[midway, above] (mag) {\Large{ $\lvert \cdot \rvert$} };
    \draw [path pic=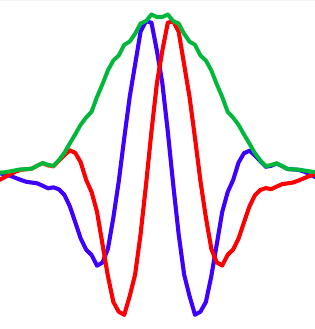, draw=white] (9.25,0,0) rectangle (11.75,2,0);
    \draw (9.25,0,0) -- (mag.north);
    \draw (11.75,0,0) -- (mag.north);
    \draw[->, fill=gray!30, ultra thick] (8.5, -2, 10) -- (10, -3, 3);

    \tikzcuboid{
    shiftx=11.5cm,
    shifty=-2cm,
    scale=0.5,
    dimx=2, dimy=2, dimz=12,
    densityx=2, densityy=2, densityz=2,
    dimzval=6C_l,
    drawxdims=false,
    drawydims=false,
    front/.style={draw=blue!50!white,fill=blue!50!white},%
    right/.style={draw=blue!50!white,fill=blue!50!white},%
    top/.style={draw=blue!50!white,fill=blue!50!white},%
    }
    \tikzcuboid{
    shiftx=11.5cm,
    shifty=-2cm,
    shiftz=8,
    scale=0.5,
    drawxdims=true,
    drawydims=true,
    dimx=2, dimy=2, dimz=4,
    densityx=2, densityy=2, densityz=2,
    dimzval=C_l,
    front/.style={draw=blue!50!white,fill=blue!50!white},%
    right/.style={draw=blue!50!white,fill=blue!50!white},%
    top/.style={draw=blue!50!white,fill=blue!50!white},%
    }
    \draw (13.2, 0.8, 0) node {\large{$z^{(l+1)}$}};
    \draw (20.5, 0.3, 0) node {\large{$x^{(l+1)}$}};
    \draw (14, -1.5, 0) node {\Large{$\conv$}};

    \tikzcuboid{
    shiftx=15.2cm,
    shifty=-1.0cm,
    shiftz=0,
    scale=0.5,
    dimx=0.4, dimy=0.4, dimz=14,
    densityx=5, densityy=5, densityz=2,
    drawxdims=false,
    drawydims=false,
    drawzdims=false,
    front/.style={draw=red!50!white,fill=red!50!white},%
    right/.style={draw=red!50!white,fill=red!50!white},%
    top/.style={draw=red!50!white,fill=red!50!white},%
    }
    \tikzcuboid{
    shifty=-1.65cm,
    }
    \tikzcuboid{
    shifty=-3.0cm,
    scale=0.5,
    drawxdims=true,
    dimxval=1,
    drawydims=true,
    dimyval=1,
    drawzdims=true,
    dimzval=7C_l,
    }
    \draw (15.5, -1.8, 0) node {$\vdots$};
    \draw [<->] (15.7, -0.8, -3) -- (15.7, -3, -3) node[near start, right] {$C_{l+1}$};
    \draw [->, fill=gray!30,ultra thick] (17.5, -1.5, 0) -- (18.5, -1.5, 0)
      node[midway, above] {$\sigma$};

    \tikzcuboid{
    shiftx=19.5cm,
    shifty=-2.0cm,
    scale=0.5,
    dimx=2, dimy=2, dimz=6,
    densityx=4, densityy=4, densityz=2,
    drawzdims=true,
    dimzval=C_{l+1},
    drawxdims=true,
    dimxval=\frac{W}{2},
    drawydims=true,
    dimyval=\frac{H}{2},
    front/.style={draw=blue!50!white,fill=blue!50!white},%
    right/.style={draw=blue!50!white,fill=blue!50!white},%
    top/.style={draw=blue!50!white,fill=blue!50!white},%
    }

  \end{tikzpicture}

  }
  \caption{Block Diagram of Proposed Invariant Layer for $J=1$. Activations are shaded
  blue, fixed parameters yellow and learned parameters red. Input
  $x^{(l)}\in \mathbb{R}^{C_l\x H\x W}$ is filtered by real and imaginary oriented
  wavelets and a lowpass filter and is downsampled. The channel dimension
  increases from $C_l$ to $(2K+1)C_l$, where the number of orientations is $K=6$.
  The real and imaginary parts are combined by taking their magnitude (an
  example of what this looks like in 1D is shown above the magnitude operator) -
  the components oscillating in quadrature are combined to give $z^{(l+1)}$. The
  resulting activations are concatenated with the lowpass filtered activations,
  mixed across the channel dimension, and then passed through a nonlinearity
  $\sigma$ to give $x^{(l+1)}$.  If the desired output spatial size is $H\x W$,
  $x^{(l+1)}$ can be bilinearly upsampled paying only a few multiplies per
  pixel.}
  \label{fig:block_diagram}
  \vspace{-10pt}
\end{figure*}
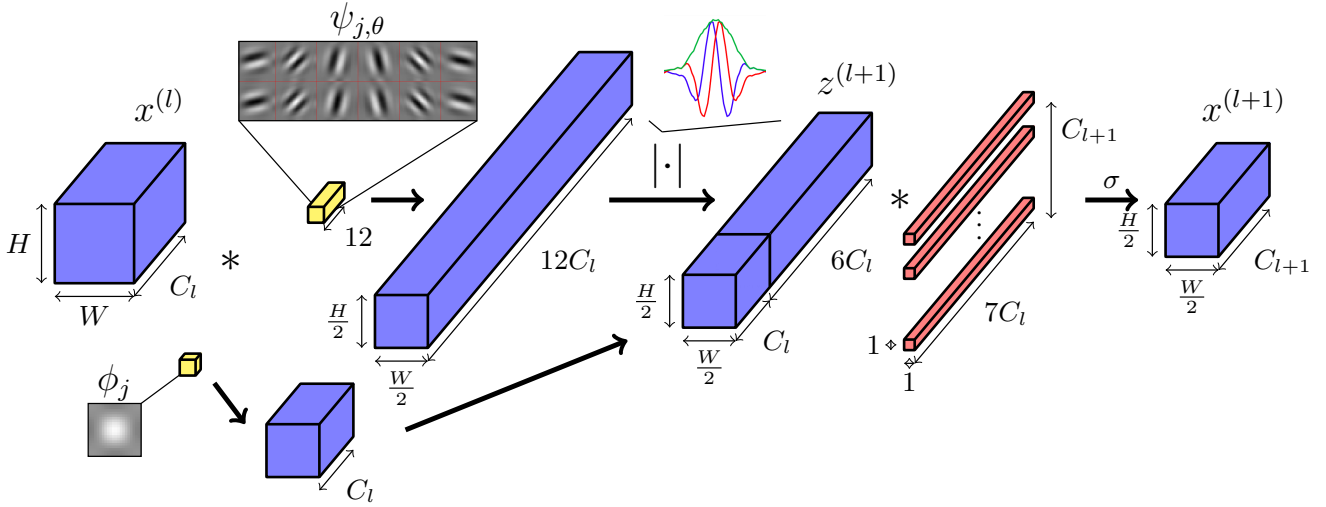

\section{Implementation}\label{sec:implementation}
Like \cite{singh_dual-tree_2017,singh_multi-resolution_2016} we use the \DTCWT
\cite{selesnick_dual-tree_2005} for our wavelet filters $\psi_{j, \theta}$ due
to their fast implementation with separable convolutions which we will discuss
more in \autoref{sec:computation}.  A side effect of this choice is that the
number of orientations of wavelets is restricted to $K=6$.

The output of the \DTCWT is decimated by a factor of $2^j$ in each direction for
each scale $j$.  In all our experiments we set $J=1$ for each invariant layer,
meaning we can mix the lowpass and bandpass coefficients at the same resolution.
\autoref{fig:block_diagram} shows how this is done. Note that setting $J=1$ for
a single layer does not restrict us from having $J>1$ for the entire system, as
if we have a second layer with $J=1$ after the first, including downsampling
($\downarrow$), we would have:
\begin{equation}
  \left(\left(\left(x \conv \phi_1\right) \downarrow 2\right) \conv \psi_{1, \theta}\right) \downarrow 2 = \left(x \conv \psi_{2, \theta}\right) \downarrow 4
\end{equation}

\subsection{Memory Cost}\label{sec:memory}
A standard convolutional layer with $C_l$ input channels, $C_{l+1}$ output channels
and kernel size $L$ has $L^2C_{l}C_{l+1}$ parameters. 

The number of learnable parameters in each of our proposed invariant layers with
$J=1$ and $K=6$ orientations is:
\begin{equation}
  \text{\#params} = (JK+1)C_{l}C_{l+1} = 7C_{l}C_{l+1}
\end{equation} 
The spatial support of the wavelet filters is typically $5\x 5$ pixels or more,
and we have reduced $\text{\#params}$ to less than $3\x3$ per filter, while
producing filters that are significantly larger than this.

\subsection{Computational Cost}\label{sec:computation}
A standard convolutional layer with kernel size $L$ needs $L^2C_{l+1}$
multiplies per input pixel (of which there are $C_{l}\x H\x W$).

As mentioned in \autoref{sec:memory}, we use the \DTCWT for our complex, shift
invariant wavelet decomposition. We use the open source Pytorch implementation
of the \DTCWT \cite{cotter_pytorch_2018} as it can run on GPUs and
has support for backpropagating gradients.

There is an overhead in doing the wavelet decomposition for each input channel. A
regular discrete wavelet transform (DWT) with filters of length $L$ will have
$2L\left(1-2^{-2J}\right)$ multiplies for a $J$ scale decomposition. A \DTCWT
has 4 DWTs for a 2-D input, so its cost is $8L\left(1-2^{-2J}\right)$, with
$L=6$ a common size for the filters. It is important to note that unlike the
filtering operation, this does not scale with $C_{l+1}$, the end result being that as
$C_{l+1}$ grows, the cost of $C_l$ forward transforms is outweighed by that of the mixing
process.

Because we are using a decimated wavelet decomposition, the sample rate decreases after each
wavelet layer. The benefit of this is that the mixing process then only works on
one quarter the spatial size after one first scale and one sixteenth the spatial
after the second scale. Restricting ourselves to $J=1$ as we mentioned in
\autoref{sec:implementation}, the computational cost is then:

\vspace{-5pt}
\begin{equation}
  \underbrace{ \frac{7}{4}C_{l+1} }_{\textrm{mixing}} +
  \underbrace{\vphantom{\frac{7}{4}} 36}_{\textrm{\DTCWT}} \quad
  \textrm{multiplies per input pixel}\label{eq:comp}
\end{equation}
In most CNNs, $C_{l+1}$ is several dozen if not several
hundred, which makes \autoref{eq:comp} significantly smaller than
$L^2C_{l+1}=9C_{l+1}$ multiplies for $3\x 3$ convolutions.

\section{Experiments}\label{sec:experiments}
In this section we examine the effectiveness of our invariant layer by testing
its performance on the well known datasets CIFAR-10 (10 classes, 5000 images per class
at $32\x 32$ pixels per image), CIFAR-100 (100 classes, 500 images per class at 
$32\x 32$ pixels per image) and Tiny ImageNet\cite{li_tiny_nodate} (a dataset
like ImageNet with 200 classes and 500 training images per class, each image at
$64 \x 64$ pixels). Our experiment code is available at
\url{https://github.com/fbcotter/invariant_convolution}.

\subsection{Layer Level Comparison}\label{sec:conv_exp}
To compare our proposed locally invariant layer (inv) to a regular convolutional
layer (conv), we use a simple yet powerful VGG-like architecture with 6 convolutional layers for CIFAR
and 8 layers for Tiny ImageNet as shown in \autoref{tab:arch}. The initial number of
channels $C$ we use is 64. Despite this simple design, this reference
architecture achieves competitive performance for the three datasets.

This network is optimized with stochastic gradient descent with momentum. The
initial learning rate is $0.5$, momentum is $0.85$, batch size $N=128$ and
weight decay is $10^{-4}$. For CIFAR-10/CIFAR-100 we scale the learning rate by
a factor of 0.2 after 60, 80 and 100 epochs, training for 120 epochs in total.
For Tiny ImageNet, the rate change is at 18, 30 and 40 epochs (training for 45 in total).

\begin{table}[ht]
  \footnotesize
    \centering
    \caption{Base Architecture used for Convolution Layer comparison tests.
    \textsuperscript{\textdagger}indicates only used in Tiny ImageNet
    experiments.}
  \label{tab:arch}
  \begin{tabular}{c|c}
    \textbf{Name} & \textbf{Output Size}  \\\hline
    convA & $C\x H\x W$ \\
    convB & $C\x H\x W$ \\
    convC & $2C\x H/2\x W/2$\\
    convD & $2C\x H/2\x W/2$\\
    convE & $4C\x H/4\x W/4$\\
    convF & $4C\x H/4\x W/4$\\
    convG\textsuperscript{\textdagger} & $8C\x H/8\x W/8$\\
    convH\textsuperscript{\textdagger} & $8C\x H/8\x W/8$\\
    fc & num classes \\
  \end{tabular}
\end{table}

We perform an ablation study where we progressively swap out convolutional
layers for invariant layers keeping the input and output activation sizes the
same. As there are 6 layers (or 8 for Tiny ImageNet), there are too many
permutations to list the results for swapping out all layers for our locally
invariant layer, so we restrict our results to swapping 1 or 2 layers. 
\autoref{tab:conv_results} reports the top-1 classification accuracies for
CIFAR-10, CIFAR-100 and Tiny ImageNet. In addition to testing on the full
datasets we report results for a reduced training set size. In the table, `invX'
means that the `convX' layer from \autoref{tab:arch} is replaced with an 
invariant layer.

Interestingly, we see improvements when one or two invariant layers are used near the
start of a system, but not for the first layer. 
\begin{table}
  \footnotesize
  \centering
  \caption{Results for testing VGG like architecture with convolutional and
  invariant layers on several datasets. An architecture with `invX' means the
  equivalent convolutional layer `convX' from \autoref{tab:arch} was swapped for
  our proposed layer.}
  \begin{tabular}{c|cc|cc|cc}
    & \multicolumn{2}{c|}{CIFAR-10} & \multicolumn{2}{c|}{CIFAR-100} & \multicolumn{2}{c}{Tiny ImgNet} \\ \cline{2-7}
    Trainset Size& 10k   &  50k  &  10k  &  50k & 20k & 100k \\ \hline
    ref & 84.4  & 91.9  & 53.2  & 70.3 & 37.4 & 59.1 \\ \hline
    invA & 82.8 & 91.3 & 48.4 & 69.5 & 33.2 & 57.7 \\
    invB & 84.8 & 91.8  & 54.8  & 70.7 & 36.9& 59.5 \\
    invC & 85.3 & 92.3 & 54.2 & 71.2 & 37.1& 59.8\\
    invD& 83.8 & 91.2 & 51.2 & 70.1 & 37.6 & 59.3 \\
    invE& 83.3 & 91.6 & 50.3 & 70.0 & 37.8 & 59.4\\
    invF& 82.1 & 90.5& 47.6& 68.9 & 34.0 & 57.8\\ \hline
    invA, invB& 83.1& 90.5 & 49.8 & 68.4 & 35.0 & 57.9\\
    invB, invC& 83.8& 91.2 & 50.6 & 69.1 & 34.6 & 57.5\\
    invC, invD& 85.1 & 92.1 & 54.3 & 70.1 & \textbf{37.9} & 59.0\\
    invD, invE& 80.2 & 89.1 & 49.0 & 67.3 & 33.9 & 57.5\\
    invA, invC& 82.8 & 90.7 & 49.5 & 69.6 & 34.0 & 56.9\\
    invB, invD& \textbf{85.4} & \textbf{92.7} & \textbf{54.6} & \textbf{71.3} & \textbf{37.9} & 59.8\\
    invC, invE& 84.8 & 91.8 & 53.5 & 70.9 & 37.6 & \textbf{60.2}\\
  \end{tabular}\label{tab:conv_results}
  \vspace{-10pt}
\end{table}

\vspace{-5pt}
\subsection{Network Comparison}\label{sec:scat_exp}
In the previous section, we examined how the locally invariant layer performs when
directly swapped in for a convolutional layer in an architecture designed
for convolutional layers. In this section, we look at how
it performs in a Hybrid ScatterNet-like \cite{oyallon_hybrid_2017,oyallon_scaling_2017},
network.

To build the second order ScatterNet, we stack two invariant layers on top of each
other. For 3 input channels, the output of these layers has $3(1 +
6 + 6 +36) = 147$ channels at $1/16$ the spatial input size. We then use 4
convolutional layers, similar to convC to convF in \autoref{tab:arch} with
$C=96$. In addition, we use dropout after these later convolutional layers with drop probability $p=0.3$.

We compare a ScatterNet with no learning in between scattering orders
(ScatNet A) to one with our proposal for a learned mixing matrix $A$ (ScatNet B). Finally,
we also test the hypothesis seen from \autoref{sec:conv_exp} about putting conv
layers before an inv layer, and test a version with a small convolutional layer
before ScatNets A and B, taking the input from 3 to 16 channels, and call these ScatNet
architectures ScatNet C and D respectively.

See \autoref{tab:hybrid_scat} for results from these experiments. It is clear from
the improvements that the mixing layer helps the Scattering front end.
Interestingly, ScatNet C and D further improve on the A and B versions
(albeit with a larger parameter and multiply cost than the mixing operation). This reaffirms that there
may be benefit to add learning before as well as inside the ScatterNet.

For comparison, we have also listed the performance of other architectures as
reported by their authors in order of increasing complexity. Our proposed ScatNet D achieves
comparable performance with the the All Conv, VGG16 and FitNet architectures.
The Deep\cite{he_identity_2016} and Wide\cite{zagoruyko_wide_2016}
ResNets perform best, but with very many more
multiplies, parameters and layers.

ScatNets A to D with 6 layers like convC to convG from \autoref{tab:arch} after
the scattering, achieve $58.1, 59.6, 60.8$ and $62.1\%$ top-1 accuracy on Tiny ImageNet. As
these have more parameters and multiplies from the extra layers we exclude them
from \autoref{tab:hybrid_scat}.
\begin{table}
  \footnotesize
  \caption{Hybrid ScatterNet top-1 classification accuracies on CIFAR-10 and CIFAR-100. 
  $N_l$ is the number of learned convolutional layers, \#param is the number of
  parameters, and \#mults is the number of multiplies per $32\x 32\x 3$ image. An asterisk indicates
  that the value was estimated from the architecture description.}\label{tab:hybrid_scat}
  \hspace{-10pt}
  \begin{tabular}{c|ccc|ccc}
    & $N_l$ &\#param&\#mults&CIFAR-10&CIFAR-100\\ \hline 
    ScatNet A & 4 & 2.6M & 165M & 89.4 & 67.0 \\ 
    ScatNet B & 6 & 2.7M & 167M & 91.1  & 70.7 \\ %
    ScatNet C & 5 & 3.7M & 251M & 91.6 & 70.8  \\ %
    ScatNet D & 7 & 4.3M & 294M & 93.0 & 73.5  \\\hline %
    All Conv\cite{springenberg_striving_2014-3} & 8 & 1.4M & 281M\textsuperscript{*} & 92.8 & 66.3 \\ %
    VGG16\cite{liu_very_2015} & 16 & 138M\textsuperscript{*} & 313M\textsuperscript{*}  & 91.6 & -  \\ 
    FitNet\cite{romero_fitnets:_2014} & 19 & 2.5M & 382M & 91.6 & 65.0 \\ %
    ResNet-1001\cite{he_identity_2016} & 1000 & 10.2M & 4453M\textsuperscript{*}& 95.1 & 77.3 \\ %
    WRN-28-10\cite{zagoruyko_wide_2016} & 28 & 36.5M & 5900M\textsuperscript{*} & 96.1 & 81.2 \\ %
  \end{tabular}
  \vspace{-15pt}
\end{table}

\section{Conclusion}\label{sec:conclusion}
In this work we have proposed a new learnable scattering layer, dubbed the
locally invariant convolutional layer, tying together ScatterNets and CNNs.
We do this by adding a mixing between the layers of ScatterNet allowing the
learning of more complex shapes than the ripples seen in
\cite{cotter_visualizing_2017}. This invariant layer can easily be shaped to allow
it to drop in the place of a convolutional layer, theoretically saving on parameters and
computation. However, care must be taken when doing this, as our ablation study
showed that the layer only improves upon regular convolution at certain depths.
Typically, it seems wise to interleave convolutional layers and invariant
layers.

We have developed a system that allows us to pass
gradients through the Scattering Transform, something that previous work has not
yet researched. Because of this, we were able to train end-to-end a system that
has a ScatterNet surrounded by convolutional layers and with our proposed mixing. 
We were surprised to see that even a small
convolutional layer before Scattering helps the network, and a
very shallow and simple Hybrid-like ScatterNet was able to achieve good
performance on CIFAR-10 and CIFAR-100.

There is still much research to do - why does the proposed layer work best near,
but not at the beginning of deeper networks? Why is it beneficial to precede an
invariant layer with a convolutional layer? Can we combine invariant layers in
Residual style architectures? The work presented here is still
nascent but we hope that it will stimulate further interest and research into both
ScatterNets and the design of convolutional layers in CNNs.

\section*{References}
\printbibliography[heading=none]
\end{document}